\documentclass[9pt,conference]{IEEEtran}
\IEEEoverridecommandlockouts
\usepackage{amsmath,amssymb,amsfonts}
\usepackage{graphicx}
\usepackage{cite}
\usepackage{booktabs}
\usepackage{stfloats}
\usepackage{verbatim,subfig}

\title{Multi-Task Learning with Additive U-Net for Image Denoising and Classification}
\author{
\IEEEauthorblockN{Vikram R. Lakkavalli\IEEEauthorrefmark{1} \\ vikram.ramesh@iiitb.ac.in, vikram.ckm@gmail.com \\Neelam Sinha\IEEEauthorrefmark{2} \\ neelam@cbr-iisc.ac.in}
\IEEEauthorblockA{\IEEEauthorrefmark{1}IIITB, Bangalore, India\\ \IEEEauthorrefmark{2}Center for Brain Research, IISc, Bangalore, India}
}

\begin{document}
\maketitle

% ------------------- ABSTRACT ------------------- %
\begin{abstract}
We investigate additive skip fusion in U-Net architectures for image denoising and denoising-centric multi-task learning (MTL). By replacing concatenative skips with gated additive fusion, the proposed Additive U-Net (AddUNet) constrains shortcut capacity while preserving fixed feature dimensionality across depth. This structural regularization induces controlled encoder–decoder information flow and stabilizes joint optimization. Across single-task denoising and joint denoising–classification settings, AddUNet achieves competitive reconstruction performance with improved training stability. In MTL, learned skip weights exhibit systematic task-aware redistribution: shallow skips favor reconstruction, while deeper features support discrimination. Notably, reconstruction remains robust even under limited classification capacity, indicating implicit task decoupling through additive fusion. These findings show that simple constraints on skip connections act as an effective architectural regularizer for stable and scalable multi-task learning without increasing model complexity.
\end{abstract}


% ------------------- INTRODUCTION ------------------- %
\section{Introduction}

Encoder--decoder architectures with skip connections are central to modern image restoration systems. In image denoising, skip connections allow the decoder to reuse encoder activations across depth, combining fine spatial detail with higher-level structure. This paradigm has also been extended to multi-task learning (MTL), where shared encoders jointly support reconstruction and recognition objectives.

In standard U-Net architectures \cite{ronneberger2015unet}, skip connections are implemented via feature concatenation. While effective, concatenation increases channel dimensionality and implicitly expands representational capacity along shortcut pathways. This expanded capacity can encourage shortcut reliance and weaken representational discipline, particularly in multi-task settings where reconstruction and recognition objectives compete for shared features. Despite the widespread use of skip connections, their role is typically treated as architectural convention rather than as a controllable design variable.

Image denoising offers a controlled setting to examine skip fusion design. Effective denoising requires suppression of stochastic corruption while preserving structurally meaningful cues such as edges and textures---features that are simultaneously relevant for recognition. This makes denoising-centric MTL a natural testbed for studying how skip design influences shared representation learning.

In this work, we adopt a structural perspective: skip connections are explicit feature reuse pathways whose capacity should be constrained rather than implicitly expanded. We propose Additive U-Net (AddUNet), which replaces concatenative skip connections with additive fusion and optional nonnegative scalar gating. This formulation preserves fixed feature dimensionality across depth and limits shortcut capacity, acting as an architectural regularizer.

We evaluate AddUNet in both single-task denoising and denoising-centric multi-task learning settings. Our analysis focuses not only on reconstruction accuracy, but also on optimization stability, generalization behavior, and the redistribution of information flow across depth.

Our contributions are as follows:
\begin{itemize}
    \item We show that additive skip fusion acts as an effective architectural regularizer in denoising-centric multi-task learning, improving optimization stability and reducing overfitting relative to concatenative U-Net designs.
    \item We demonstrate that AddUNet achieves denoising performance competitive with strong CNN-based baselines despite operating with fixed feature dimensionality and constrained shortcut capacity.
    \item We provide empirical evidence of task-aware skip-weight reorganization in MTL, revealing an emergent separation between shallow reconstruction-dominant features and deeper discriminative representations.
\end{itemize}

% ------------------- RELATED WORK ------------------- %
\section{Related Work}

\subsection{Image denoising}
Residual CNNs such as DnCNN \cite{zhang2017dncnn} established strong baselines for Gaussian denoising, with later extensions incorporating blind-noise training and non-local operations \cite{zhang2018ffdnet,liu2018nonlocal}. Encoder--decoder architectures such as U-Net \cite{ronneberger2015unet} remain popular due to their ability to recover fine detail through skip connections. Most prior work emphasizes reconstruction fidelity measured by PSNR or SSIM, with limited analysis of how skip fusion design affects optimization and generalization.

\subsection{Skip connection designs}
Concatenative skip connections increase representational capacity by expanding channel dimensionality. Alternative designs include residual shortcuts \cite{he2016resnet}, attention-based modulation \cite{oktay2018attention,woo2018cbam}, and conditional feature modulation \cite{perez2018film}. While these approaches increase flexibility or spatial selectivity, skip connections are often treated as implicit feature aggregators rather than as architectural choices that directly constrain feature reuse capacity.

\subsection{Multi-task learning}
Multi-task learning has been widely studied as a means of improving generalization by sharing representations across related tasks \cite{caruana1997multitask,kendall2018multi}. In vision, MTL frameworks commonly share encoder representations with task-specific heads. However, relatively little attention has been paid to how skip connection design influences the stability of shared representations when reconstruction and recognition objectives are optimized jointly.

\noindent In contrast to prior work, we focus on constraining skip fusion itself through additive formulation, demonstrating that such architectural regularization improves stability and generalization in denoising-centric multi-task learning without introducing attention mechanisms or increasing model capacity.

% ------------------- METHOD ------------------- %
\section{Proposed Method: Additive U-Net}
\label{sec:method}

We adopt a U-Net--based encoder--decoder architecture with additive skip fusion. The design enforces three constraints:
(i) a fixed feature dimensionality across all encoder and decoder stages;
(ii) additive skip fusion in place of concatenation; and
(iii) optional nonnegative scalar gates that modulate skip contribution during reconstruction.

These constraints limit unrestricted feature reuse and avoid channel inflation, yielding a lightweight architecture with stable optimization behavior.

\begin{figure}[ht]
\centering
\includegraphics[scale=0.5]{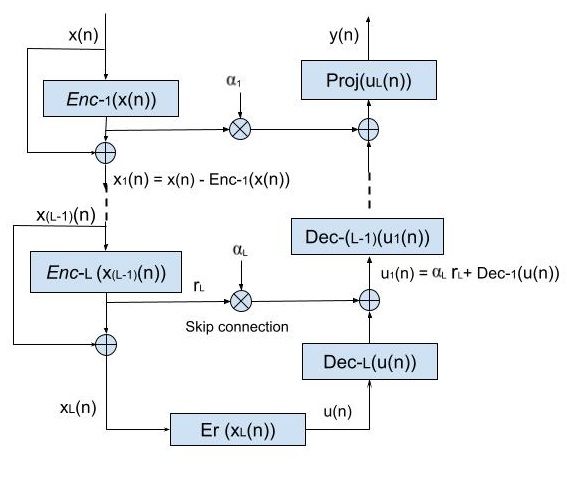}
\caption{\textit{Additive U-Net with scalar-gated additive skip fusion. Encoder and decoder operate in a shared feature space with fixed dimensionality across depth.}}
\label{fig:unetArchitecture}
\end{figure}

% ------------------- LOSS ------------------- %
\subsection{Loss and Training}
\label{sec:loss}

AddUNet is trained in a multi-task learning setting with two objectives: image denoising and image classification. Given a noisy image $x$ and clean target $y$, the network outputs a denoised estimate $\hat{y}$ and a classification prediction $\hat{p}$ from a lightweight auxiliary head attached to the encoder.

\paragraph{Denoising loss.}
We use the Charbonnier loss:
\begin{equation}
\mathcal{L}_{\mathrm{den}} = \sqrt{(y - \hat{y})^2 + \epsilon^2}, \quad \epsilon = 10^{-3}.
\end{equation}

\paragraph{Classification loss.}
Cross-entropy loss is used for classification:
\begin{equation}
\mathcal{L}_{\mathrm{cls}} = -\log \hat{p}(c).
\end{equation}

\paragraph{Joint objective.}
\begin{equation}
\mathcal{L} = \mathcal{L}_{\mathrm{den}} + \lambda \mathcal{L}_{\mathrm{cls}}.
\end{equation}

% ------------------- EXPERIMENTS ------------------- %
\section{Experimental Setup}

We evaluate the proposed AddUNet in both single-task denoising and denoising-centric multi-task learning (MTL) settings. In all experiments, denoising is treated as the primary task, while classification serves as an auxiliary objective that regularizes the shared encoder representation. Reconstruction quality is evaluated using PSNR and SSIM, and recognition performance is evaluated using classification accuracy.

\subsection{Datasets and noise models}

Single-task denoising experiments are conducted on the Kodak-17 grayscale dataset using additive white Gaussian noise (AWGN) with standard deviations $\sigma \in \{15,25,50\}$, following conventional practice for 8-bit images prior to normalization.

Denoising-centric MTL experiments are performed on MNIST, Fashion-MNIST, EMNIST (balanced split), and STL-10. For all MTL datasets, input images are normalized to the $[0,1]$ range, and noise parameters are defined directly in the normalized domain. Gaussian noise with $\sigma = 0.2$ is used for MNIST, Fashion-MNIST, and EMNIST, optionally combined with salt-and-pepper noise with probability $p = 0.1$ in selected experiments. STL-10 experiments follow the same normalized noise convention. This ensures that reconstruction and classification losses operate on compatible signal scales in MTL settings, at the same time preserving comparability with the established denoising benchmarks.

\subsection{Training configurations}

In the denoise-only setting, AddUNet is trained solely with the reconstruction objective. In the MTL setting, a lightweight classification head is attached to the encoder bottleneck, and the network is trained jointly using a weighted sum of denoising and classification losses. The classification head is intentionally kept simple to isolate the effect of additive skip fusion on shared representation learning rather than classifier capacity.

The weighting coefficient $\lambda$ controlling the contribution of the classification loss is ramped during training, allowing the network to first stabilize reconstruction before gradually introducing discriminative pressure. This schedule is shared across all MTL experiments.

\subsection{Architectural variants and baselines}

We compare AddUNet against three classes of baselines:
(i) DnCNN, representing a strong residual CNN-based denoising model without encoder--decoder structure; and
(ii) a pseudo-additive U-Net that replaces concatenative skip connections with direct feature addition but does not employ subtractive residual encoding or gated skip fusion.
(iii) AutoEncoder (AE) based architecture which doesn't have skip paths

All AddUNet variants operate with fixed channel dimensionality across depth, ensuring that differences in performance arise from skip fusion design rather than parameter count or channel expansion.

% ------------------- RESULTS ------------------- %
\section{Results and Discussion}

\subsection{Denoising performance}
AddUNet achieves PSNR and SSIM comparable to DnCNN and pseudo-additive baselines across noise levels, despite its constrained design and fixed channel dimensionality.
\begin{table}[ht]
\centering
\caption{\textit{Average PSNR (dB) / SSIM on Kodak-17 grayscale images for Gaussian noise
$\sigma \in \{15,25,50\}$. Auto-Encoder (AE) uses $3\times3$ kernels across 5 levels with down/up sampling. P-AddU represents pseudo-additive U-Net and R-AddU denotes real Additive U-Net based denoisers}}
\setlength{\tabcolsep}{3pt}
\label{tab:psnr}

\begin{tabular}{lccc}

\toprule
Model (depth, kernels) & $\sigma{=}15$ & $\sigma{=}25$ & $\sigma{=}50$ \\
\midrule
AE (5, 3--3--3--3--3)                 & 28.44 / 0.829 & 27.55 / 0.792 & 25.85 / 0.720 \\
P-AddU (5, 3--3--3--3--3)               & \textbf{32.03} / \textbf{0.898} & \textbf{29.68} / \textbf{0.844} & \textbf{26.49} / \textbf{0.743} \\
R-AddU (5, 3--3--3--3--3)               & 31.53 / 0.888 & 29.21 / 0.833 & 25.34 / 0.709 \\
R-AddU (5, 5--5--5--5--5)               & 31.03 / 0.879 & 28.98 / 0.821 & 25.67 / 0.707 \\
R-AddU (5, 9--7--5--3--1)               & 31.73 / 0.896 & 28.88 / 0.828 & 25.23 / 0.719 \\
R-AddU (5, 1--3--5--7--9)               & 30.16 / 0.887 & 27.46 / 0.827 & 24.00 / 0.721 \\
R-AddU (3, 3--3--3)                     & 31.63 / 0.891 & 29.34 / 0.836 & 26.31 / 0.732 \\
\bottomrule
\end{tabular}
\end{table}

Table~\ref{tab:psnr} reports the average PSNR and SSIM on Kodak-17 for Gaussian noise at $\sigma \in {15,25,50}$. Across configurations, AddUNet achieves denoising performance comparable to pseudo-additive and residual baselines, while operating with fixed channel dimensionality. The results are consistent across noise levels, with performance remaining competitive in the moderate and high noise regimes. Variations in kernel scheduling and depth demonstrate that denoising performance can be adjusted within the additive framework through architectural design choices.

\begin{figure*}[ht]
\centering
\includegraphics[scale=0.38]{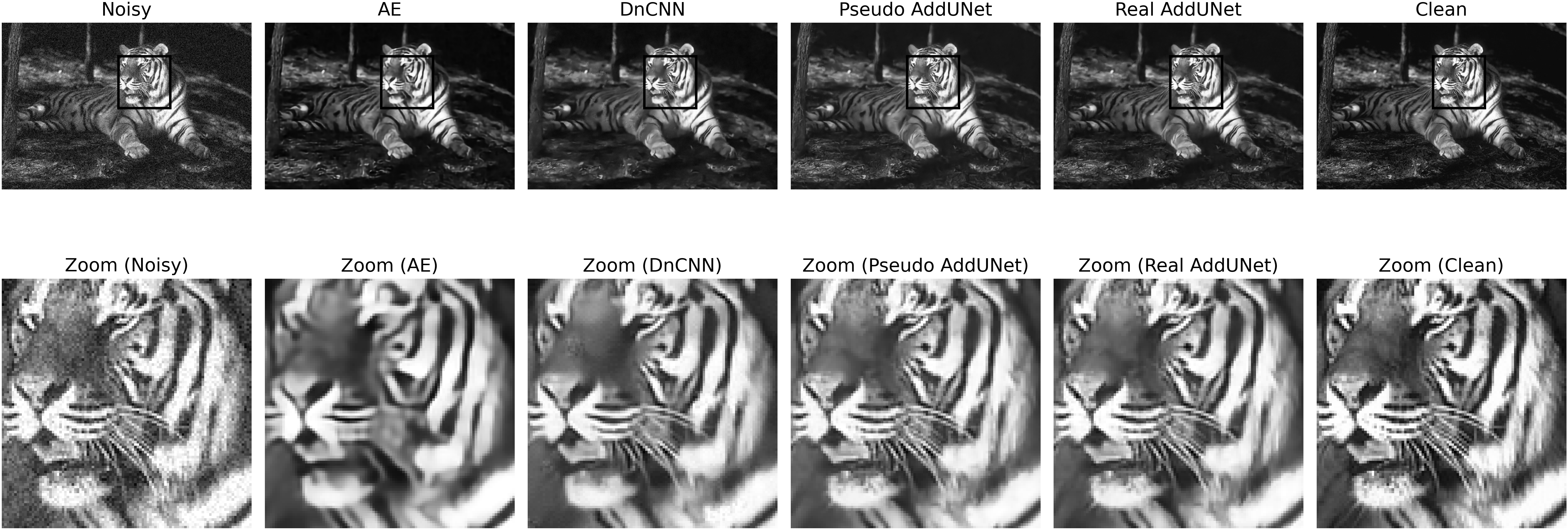}
\caption{\textit{Visual comparison of denoising methods on a Kodak-17 grayscale image with Gaussian noise ($\sigma=15$). The zoomed region highlights that the proposed Real AddUNet preserves fine structural details more faithfully than AE, DnCNN, and Pseudo AddUNet, while effectively suppressing noise.}}
\label{fig:tiger}
\end{figure*}

Figure~\ref{fig:tiger} presents a visual comparison of denoising results produced by DnCNN, pseudo-AddUNet, and real-AddUNet for a Gaussian-corrupted input with $\sigma = 15$. While P-AddUNet achieves slightly higher PSNR in Table~\ref{tab:psnr}, visual inspection reveals that real-AddUNet preserves edge continuity and boundary transitions more consistently in the highlighted regions. This suggests that quantitative gains in PSNR do not always correspond to improved structural coherence.

\begin{figure}[t]
  \centering
  \includegraphics[scale=0.28]{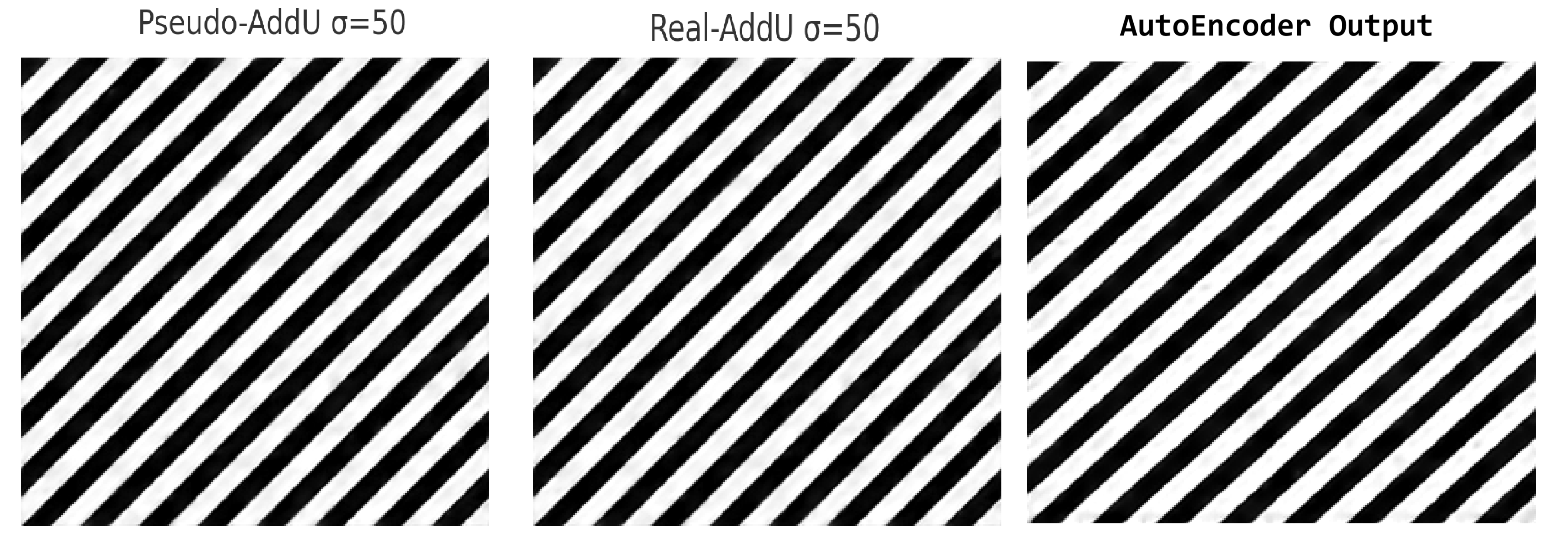}
\caption{\textit{Denoising on a structured high-frequency checker pattern at $\sigma = 50$. 
Pseudo-AddUNet achieves higher PSNR (22.11 dB) compared to Real-AddUNet (20.83 dB), 
indicating lower pixel-wise error. Real-AddUNet has higher SSIM (0.852 vs. 0.820), reflecting improved structural consistency. pseudo-AddUNet favors intensity-level fidelity, and the constrained residual in Real-AddUNet promotes more uniform spatial reconstruction.}}

  \label{fig:checker-edge}
\end{figure}

This behavior becomes more pronounced in Figure~\ref{fig:checker-edge}, where under severe corruption ($\sigma = 50$), real-AddUNet maintains sharper and more uniform stripe alignment compared to pseudo-AddUNet, indicating increased robustness in preserving structured high-frequency patterns despite comparable reconstruction metrics. 
\begin{figure}[ht]
\centering
\includegraphics[scale=0.5]{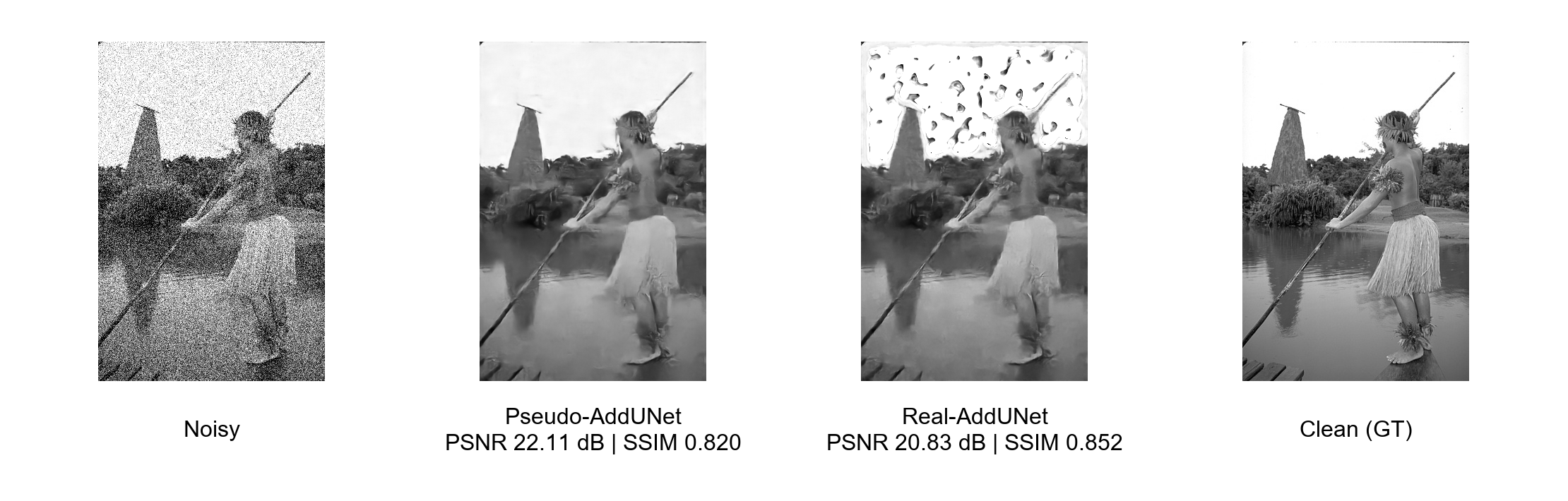}
\caption{\textit{Comparison at $\sigma=50$. Pseudo-AddUNet achieves higher PSNR (22.11 dB), whereas Real-AddUNet yields higher SSIM (0.852 vs. 0.820), suggesting better structural preservation despite lower PSNR.}}

\label{fig:pseudo_real_addUnet_compare}
\end{figure}

In images exhibiting saturation-like corruption, pseudo-AddUNet outperforms real-AddUNet in both PSNR and SSIM. For example, on Fig \ref{fig:pseudo_real_addUnet_compare}, pseudo-AddUNet improves PSNR by 4.8 dB and SSIM by 0.096 compared R-AddUnet suggesting that unconstrained skip fusion can be advantageous when aggressive intensity correction is required. In contrast, the constrained additive residual formulation prioritizes controlled feature propagation, which may limit recovery amplitude under extreme degradations.

For grayscale Kodak images, supervised denoising benchmarks \cite{zhang2024overview} typically report PSNR values of 34–35 dB at $\sigma=15$, around 30–31 dB at $\sigma=25$, and 26–27 dB at $\sigma=50$. Within this context, AddUNet achieves competitive performance at $\sigma=25$ and $\sigma=50$, while remaining below peak PSNR at lower noise levels. This behavior is consistent with prior observations that lightweight or constrained architectures trade peak fidelity for robustness and interpretability. Notably, at higher noise levels where structural preservation dominates, AddUNet performs comparably to more complex models despite its significantly simpler design.

\subsection{Effect of additive skips}

\begin{figure}[t]
\centering
\includegraphics[scale=0.25]{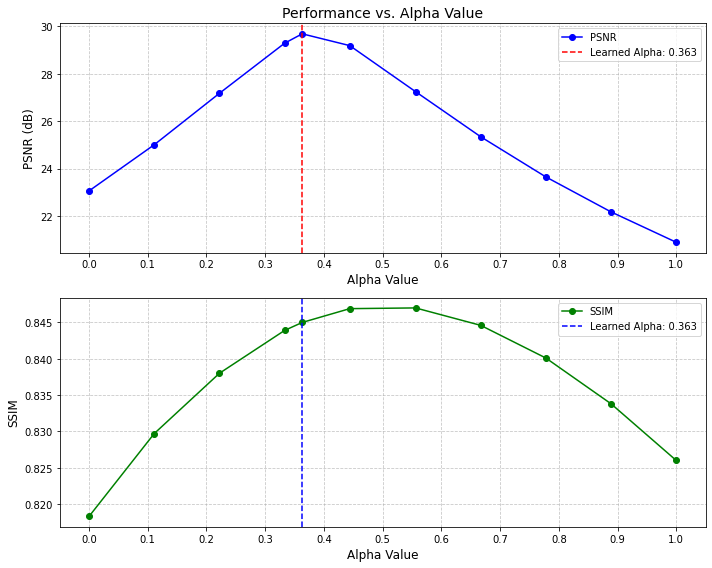}
\caption{\textit{Effect of skip-gate modulation at inference. We vary the scalar gate $\alpha_3$ in the range $[0,1]$ for a fixed input image while keeping all network parameters fixed. Using the model with a 9-7-5-3-1 kernel schedule, PSNR and SSIM vary smoothly and peak near the value learned during training. Here, $\alpha_3$ controls the additive contribution of the third-level skip connection, providing a direct post-training control knob over skip fusion.}}

\label{fig:alpha}
\end{figure}

We further examine the role of the scalar skip gate $\alpha$ in controlling reconstruction behavior. In AddUNet, each skip connection is modulated by a nonnegative scalar parameter learned during training. Importantly, this parameter can be adjusted at inference time without retraining the network, providing a direct mechanism to regulate encoder–decoder feature fusion.

Figure~\ref{fig:alpha} illustrates the effect of varying $\alpha_3$ while keeping all other network parameters fixed. The results are shown for the depth-5 model with a 9--7--5--3--1 encoder kernel schedule. As $\alpha_3$ is swept across the range $[0,1]$, PSNR and SSIM vary smoothly, peaking near the value learned during training. Crucially, modifying $\alpha_3$ does not induce instability or abrupt degradation in reconstruction quality; instead, performance changes in a controlled and predictable manner. This behavior demonstrates that additive skip fusion provides a stable and interpretable control mechanism over intermediate-scale feature contribution, even after training.

\begin{figure}[ht]
  \centering
  \includegraphics[scale=0.08]{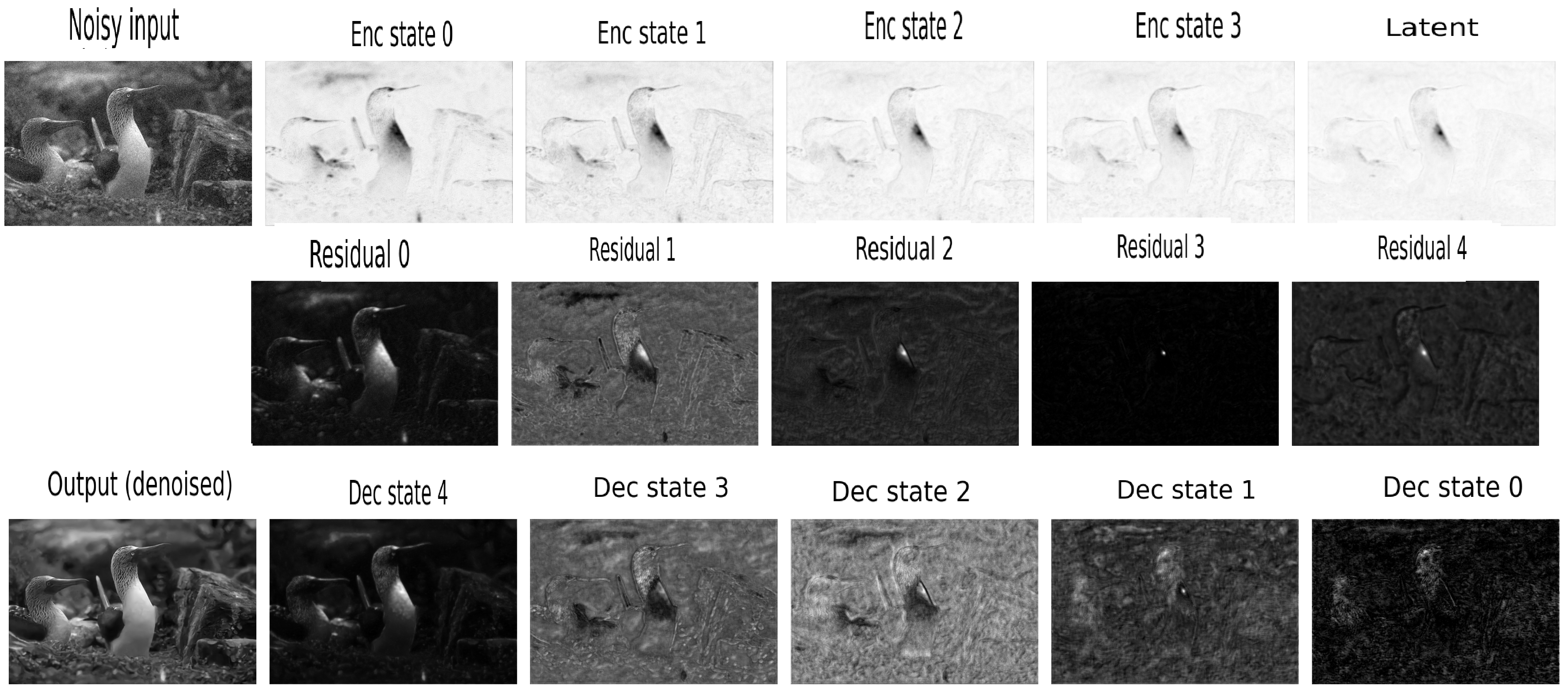}
  \caption{\textit{Visualization of intermediate representations in the Real Additive U-Net. The top row shows the noisy input and input to successive encoder states, illustrating progressive residual removal. The middle row visualizes residual components extracted at each depth. The bottom row shows decoder states and the final denoised output, demonstrating incremental reconstruction via gated additive fusion.}}

  \label{fig:kernel-comparison}
\end{figure}
\subsection{Encoder Filter Characteristics}
\begin{figure}[t]
    \centering
    \subfloat[Radial frequency profiles across encoder layers]{
        \includegraphics[width=0.25\textwidth]{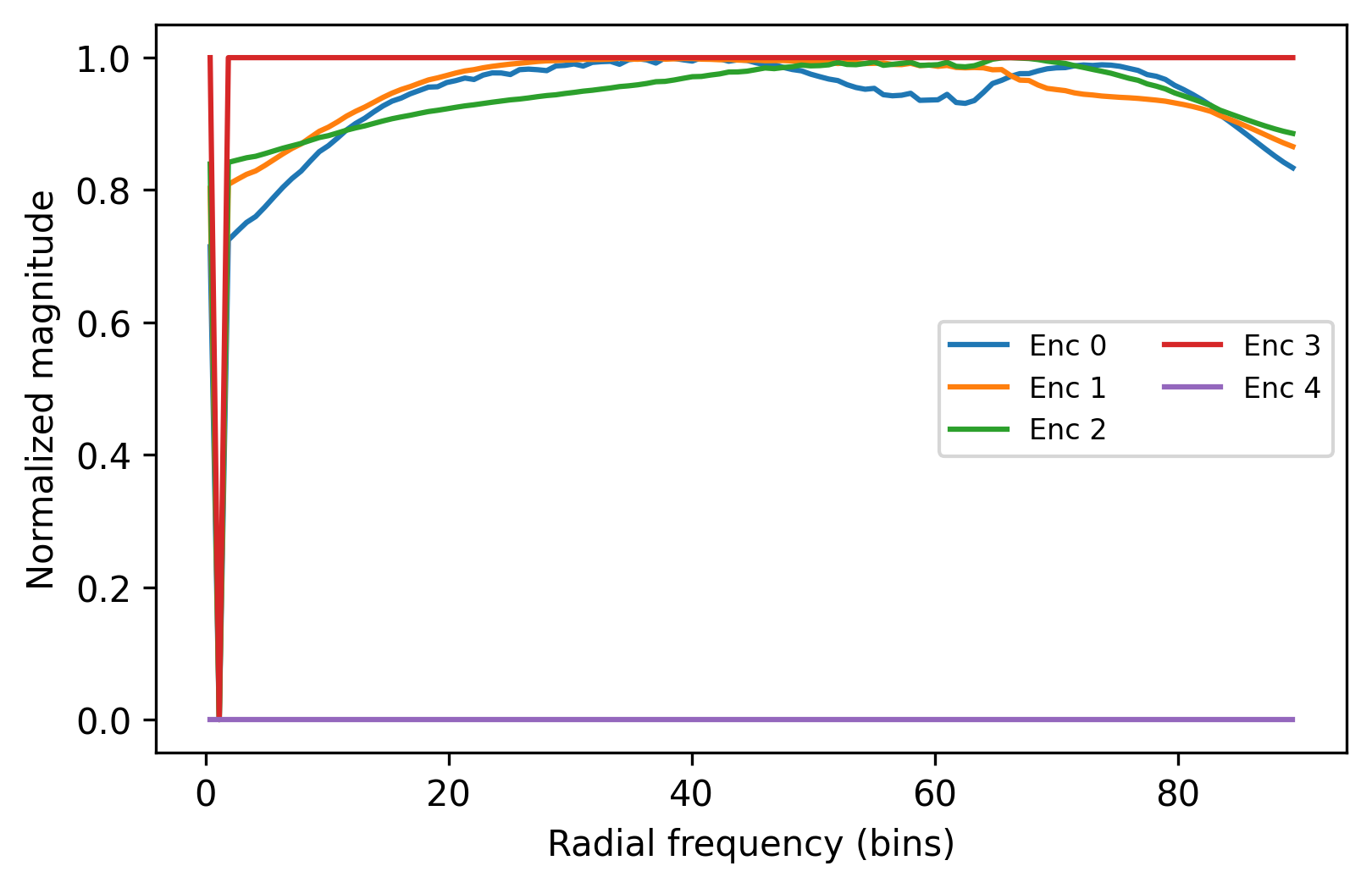}
    }
    \hfill
    \subfloat[Encoder 0: Top-$K$ filter FFT exemplars]{
        \includegraphics[width=0.2\textwidth]{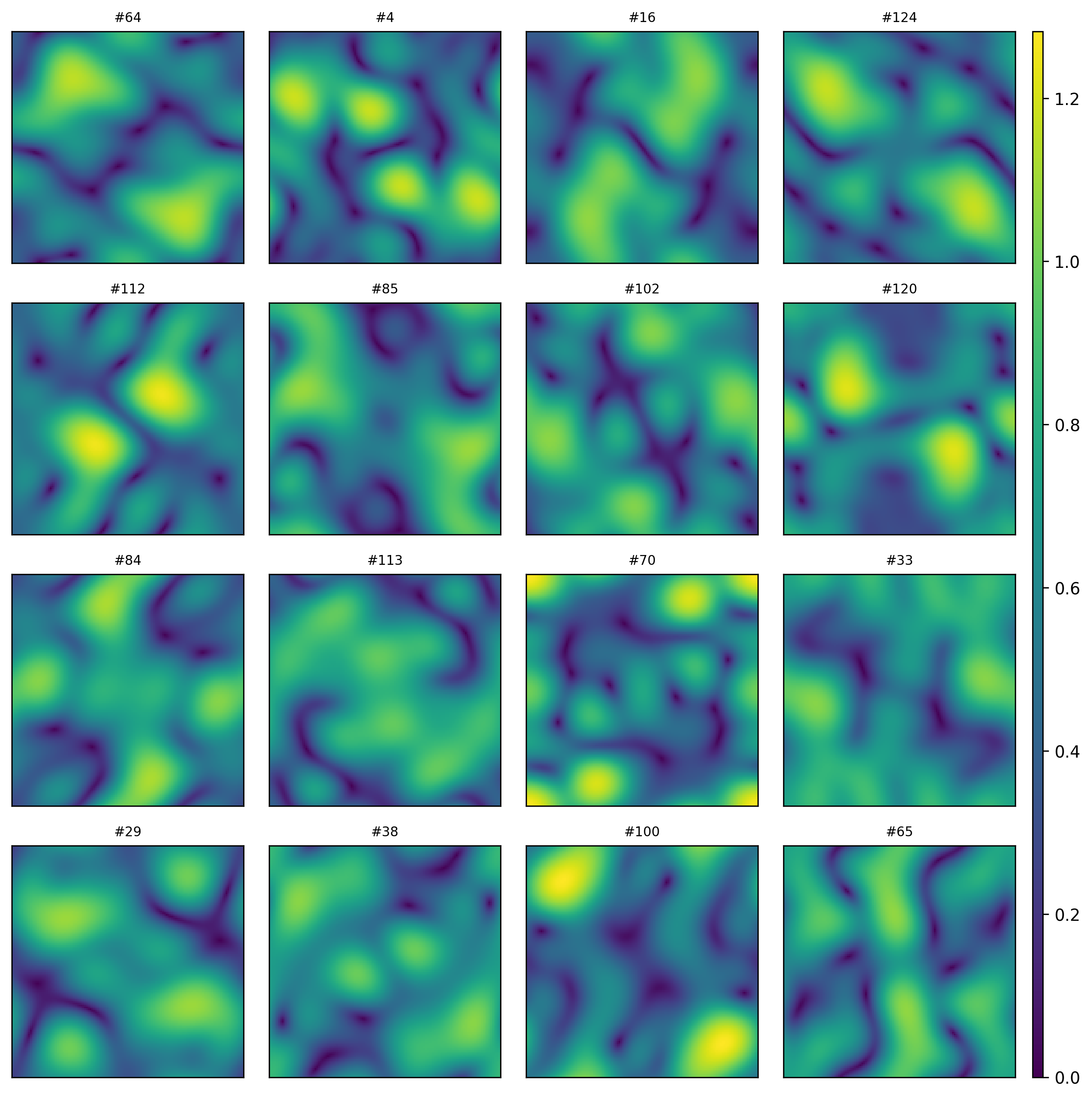}
    }
    \vspace{1em}
    \subfloat[Encoder 1: Top-$K$ filter FFT exemplars]{
        \includegraphics[width=0.2\textwidth]{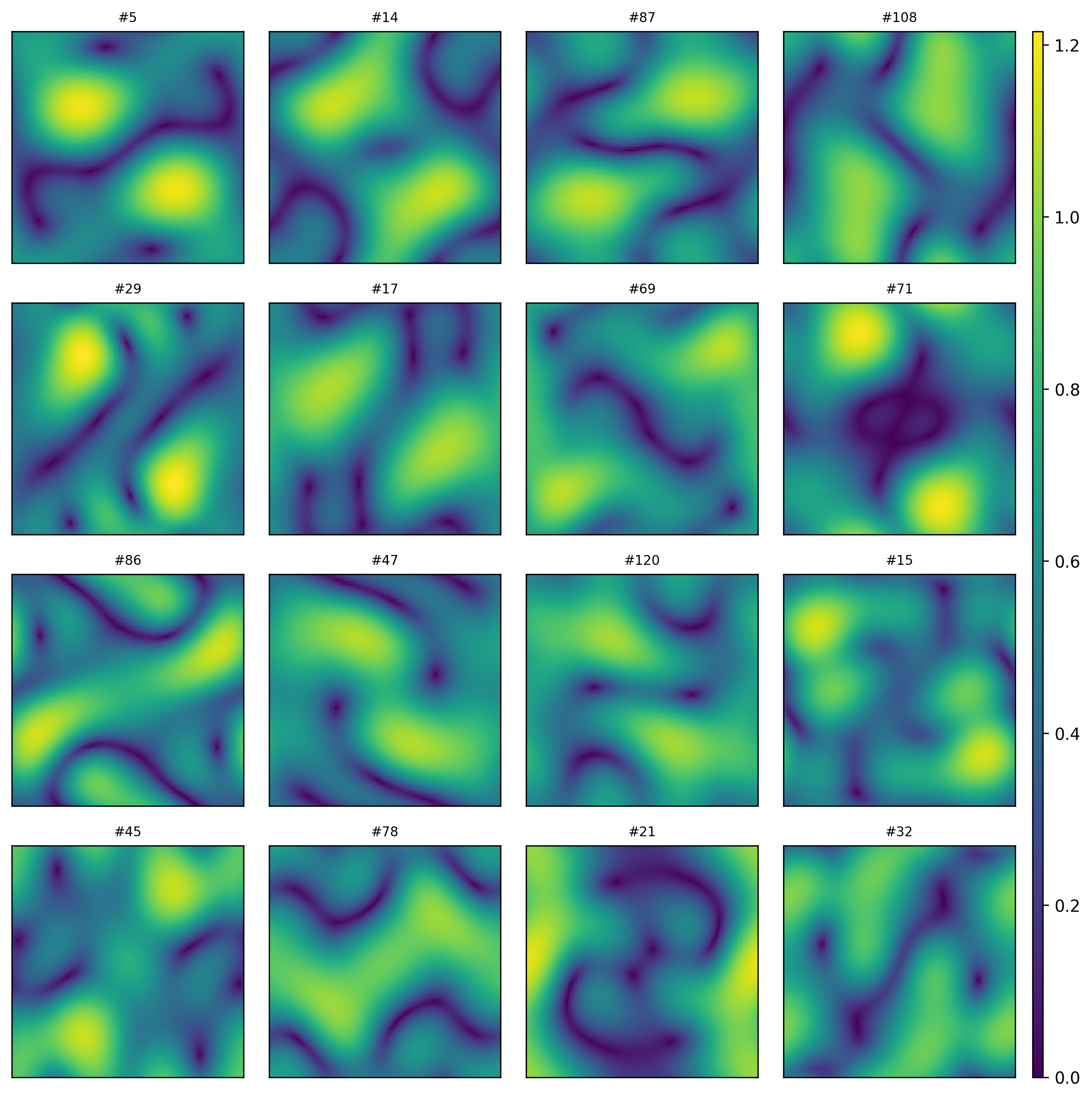}
    }
    \hfill
    \subfloat[Encoder 2: Top-$K$ filter FFT exemplars]{
        \includegraphics[width=0.2\textwidth]{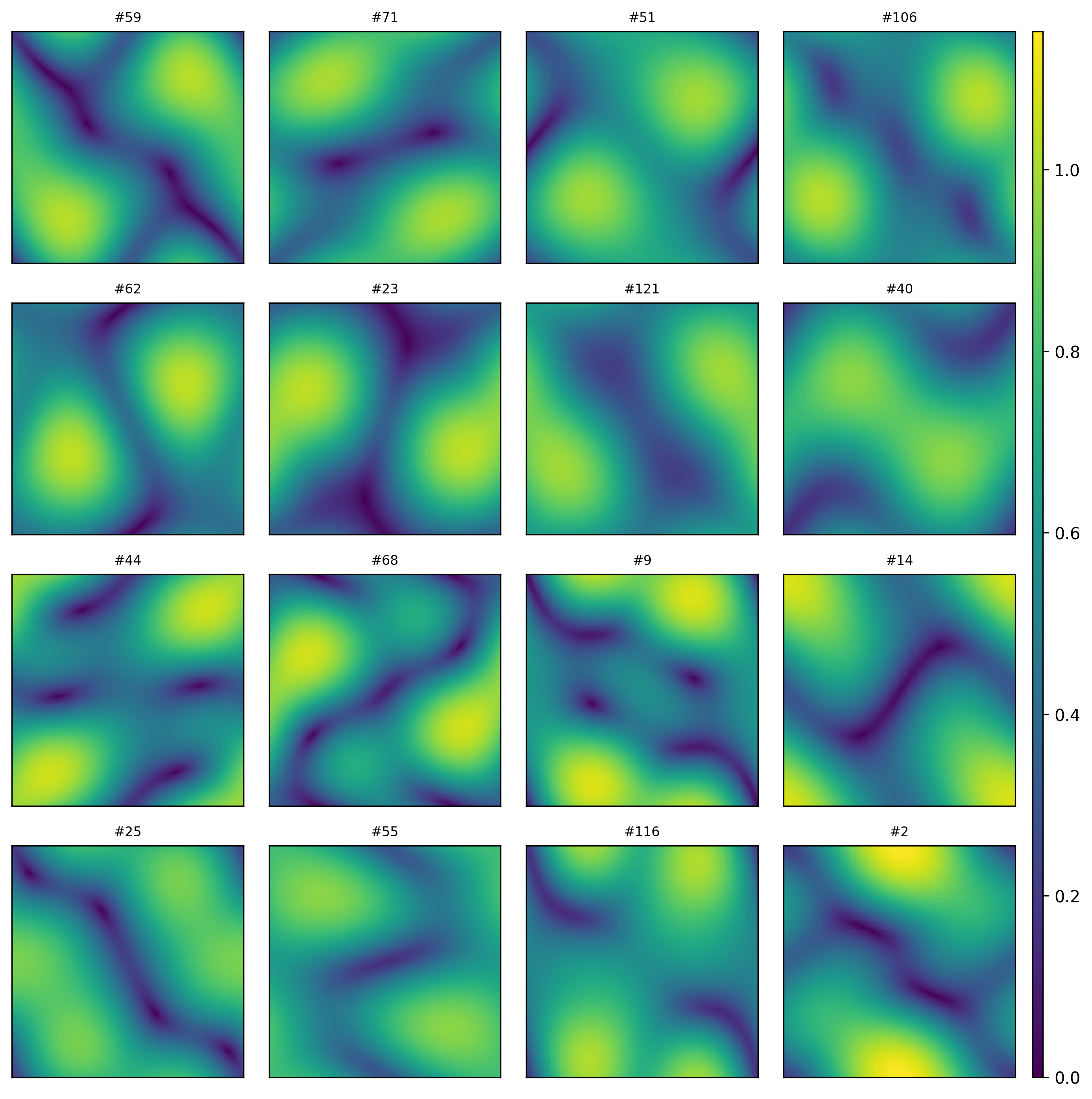}
    }
    \caption{\textit{Frequency analysis of encoder filters. 
Shallow layers emphasize high frequencies, deeper layers emphasize low 
frequencies. Thus, early skips carry fine detail and later skips convey 
coarse structure, forming an emergent frequency hierarchy.}}
    \label{fig:addunet_fft}
\end{figure}

Figure~\ref{fig:addunet_fft} presents a frequency-domain analysis of the learned encoder representations. We observe an emergent hierarchical organization, where encoder layers closer to the input exhibit stronger high-frequency responses, while deeper layers progressively emphasize lower-frequency components. As a result, skip connections originating from early encoder stages predominantly convey fine-scale details, whereas deeper skips transmit coarser structural information to the decoder. Notably, this hierarchy arises naturally from the additive residual formulation, without explicit multi-scale constraints, and mirrors the functional role of enforced scale separation in classical U-Net architectures.

\subsection{Denoising-Centric Multi-Task Learning}

We now examine whether AddUNet can simultaneously perform both denoising and recognition effectively, rather than favoring one objective at the expense of the other. Table~\ref{tab:emnist_denoise_mtl} compares denoise-only training with denoising-centric multi-task learning (MTL) on EMNIST.

In the denoise-only setting, AddUNet achieves strong reconstruction performance (26.97 dB PSNR, 0.979 SSIM), with skip weights distributed gradually across depth. When classification is introduced as an auxiliary task, reconstruction performance degrades only marginally (0.44 dB PSNR drop), while classification accuracy reaches 89.4\%. This indicates that the shared encoder can support a discriminative objective without compromising its denoising role.

Crucially, the learned skip-weight vectors reveal a systematic reorganization of information flow under MTL. In the denoise-only case, intermediate and deep skip connections retain non-negligible weights, reflecting multi-scale feature reuse for reconstruction. Under MTL, deeper skip contributions are strongly suppressed, while shallow skips dominate. This behavior suggests an implicit separation of roles: early encoder features are preferentially reused for reconstruction, while deeper representations are preserved for abstraction useful to classification.

We further evaluated AddUNet-based MTL on the more challenging STL-10 dataset. While denoising performance remains stable and competitive across training regimes, classification accuracy is notably lower than EMNIST. This behavior is expected, as STL-10 contains higher-resolution natural images with complex semantic variability, while our experiments intentionally employ a lightweight CNN-based classification head. These results highlight an important property of the proposed framework: even when the auxiliary task is under-parameterized, denoising performance does not collapse or regress. In contrast to concatenative U-Net designs, additive gated skip fusion prevents deep semantic features from leaking into the reconstruction pathway, thereby preserving denoising fidelity despite suboptimal classification capacity.

Overall, these results demonstrate that AddUNet enables coordinated multi-task optimization, where reconstruction quality is preserved and task-specific information flow is adaptively regulated through learned skip weights, rather than enforced architectural separation.

\begin{table*}[ht!]
\centering
\caption{\textit{Comparison of AddUNet performance and learned skip-weight vectors on EMNIST under denoise-only and multi-task learning (MTL) settings. Reconstruction metrics correspond to the best validation epoch. Skip weights $\boldsymbol{\alpha}=[\alpha_1,\dots,\alpha_5]$ are reported at convergence, ordered from shallow to deep encoder levels.}}
\label{tab:emnist_denoise_mtl}
\begin{tabular}{lcccc}
\toprule
Setting &
PSNR (dB) &
SSIM &
Accuracy &
$\boldsymbol{\alpha}$ (final) \\
\midrule
Denoise-only &
26.97 &
0.9794 &
-- &
$[0.259,\;0.205,\;0.205,\;0.172,\;0.160]$ \\

MTL (Denoise + Cls) &
26.53 &
0.9773 &
0.894 &
$[0.188,\;0.071,\;0.057,\;0.079,\;0.022]$ \\
\bottomrule
\end{tabular}
\end{table*}

\begin{table}[ht]
\centering
\setlength{\tabcolsep}{3pt}
\caption{\textit{Classification performance on EMNIST (balanced, 47 classes).
Mixed noise consists of Gaussian noise ($\sigma=0.2$) combined with salt-and-pepper noise ($p=0.1$).
Best test accuracy is reported along with train accuracy at the same epoch and the resulting generalization gap.}}
\label{tab:emnist_cls_gap}
\begin{tabular}{lccc}
\toprule
\textbf{Model} & \textbf{Train Acc.} & \textbf{Test Acc.} & \textbf{Gap} \\
\midrule
MLP (2-layer, 256), Clean                & 89.50 & 84.41 & 5.09 \\
MLP (2-layer, 256), Mixed Noise          & 84.97 & 81.94 & 3.03 \\
CNN (5--3--3), Clean                     & 91.78 & 88.53 & 3.25 \\
CNN (5--3--3), Mixed Noise               & 90.45 & 87.62 & 2.83 \\
\midrule
AddUNet (7--7--7--7--7) + MLP, Clean & 91.90 & 89.58 & 2.32 \\
AddUNet--MTL (7--7--7--7--7), Mixed Noise   & 91.27 & \textbf{89.32} & 1.94 \\
AddUNet--MTL (9--9--9--9--9), Mixed Noise   & 90.51 & 89.19 & \textbf{1.32} \\
\bottomrule
\end{tabular}
\end{table}

From the Table~\ref{tab:emnist_cls_gap}, it may be observed that classifier-only baselines show significant overfitting on EMNIST, while noise regularization reduces the gap at the cost of accuracy. AddUNet-MTL achieves a substantially smaller train–test gap under both clean and noisy conditions, highlighting the effectiveness of denoising-centric multi-task learning as a lightweight and principled regularizer.

% ------------------- CONCLUSION ------------------- %
\section{Conclusion}

We introduced AddUNet, an encoder–decoder architecture that replaces concatenative skip connections with additive fusion and optional scalar gating. The central contribution of this work is to treat skip connections as controllable feature reuse operators rather than passive aggregation pathways.

Additive skip fusion enforces fixed feature dimensionality and constrains shortcut capacity, preventing uncontrolled channel expansion. This structural constraint acts as an architectural regularizer, improving optimization stability and reducing overfitting in denoising-centric multi-task learning. Unlike concatenative designs, additive fusion exposes interpretable scalar gates that remain externally adjustable at inference time, enabling stable and predictable modulation of encoder–decoder information flow without retraining.

Across denoising benchmarks, AddUNet achieves performance comparable to strong CNN baselines despite reduced shortcut capacity. In multi-task settings, learned skip weights exhibit systematic task-aware redistribution, supporting implicit separation between reconstruction and discrimination.

Comparison between pseudo-additive and constrained additive formulations further reveals a trade-off between aggressive intensity correction and structural regularization, reinforcing the importance of explicitly regulating skip fusion.

These findings suggest that skip design is not merely an implementation detail but a principled architectural choice that directly governs representation flow and multi-task stability.

% ------------------- REFERENCES ------------------- %
\bibliographystyle{IEEEbib}
%\bibliography{refs}

\end{document}